# Predictive Conformal Slip Monitoring: An Empirical Evaluation of Rolling Split Conformal Prediction for Pre-Incident Traction Loss Detection

*A negative-result evaluation against ground-truth incident labels*

**Varshith Roy Kotla**

*Department of CSE, Faculty of Science and Technology, The ICFAI Foundation for Higher Education, Hyderabad, India*

*July 2026*

---

**Dataset:** 2023 Italian Grand Prix (Monza) full race-session telemetry, all laps, 19 of 20 classified drivers (one driver excluded for insufficient post-filter samples)

**Ground truth:** FIA Race Control Messages and track-limits lap deletions for the same session

---

## Abstract

Conventional traction control architectures intervene only after the adhesion limit of a tire has already been breached. This paper investigates whether Rolling Split Conformal Prediction , monitoring the volatility of non-conformity residuals from a per-driver Random Forest model of expected slip behavior , can serve as a statistically grounded pre-incident warning signal, ahead of gross traction loss. Unlike an earlier internal draft of this work, the evaluation reported here corrects a confound in the slip proxy (vehicle speed is included as an explicit model feature, not left implicit in the target's denominator), uses every racing lap for each driver rather than only the fastest lap, and is scored against real, timestamped incident labels extracted from FIA Race Control Messages and track-limits lap deletions rather than narrated post-hoc. The result is negative: across 19 drivers and 55,563 test-phase telemetry samples, the rolling-volatility detector achieves a mean precision of essentially 0.0 and mean recall of 0.0 against 14 ground-truth incidents, while flagging on average 15.3% of all samples as anomalous , too high a false-alarm rate for any early-warning use. A static 95th-percentile threshold baseline performs no better in any way that would justify the added complexity of the conformal-volatility formulation. Residual autocorrelation diagnostics show the split-conformal exchangeability assumption is violated for every driver (Ljung-Box $p < 0.001$, $n = 19/19$), which is one plausible driver of the high false-alarm rate. We report this as a methodologically rigorous negative finding, diagnose its likely causes, and outline what a genuinely predictive version of this approach would require.

---

## 1. Introduction

Traction control systems have historically operated as threshold managers: a slip ratio exceeds a defined limit, torque is cut, the tire reloads. The appeal of a pre-incident signal , some statistical quantity that rises before the physical adhesion limit is crossed , is obvious. Conformal prediction, a distribution-free framework for producing valid uncertainty estimates, seemed a natural fit: build a per-driver model of "normal" mechanical behavior, and treat rising volatility in that model's prediction errors as a marker of an approaching chaotic transition in tire behavior.

This paper tests that idea directly against real telemetry and real incident records from the 2023 Italian Grand Prix, rather than assuming it works. We do not report a successful detector. We report what happened when the idea was implemented carefully and checked against ground truth, and we treat that outcome as the actual finding of the paper.

The remainder of the paper is organized as follows: Section 2 describes the corrected theoretical framework, including the design change that removes a proxy confound present in an earlier formulation of this work. Section 3 describes how ground-truth incident timestamps were extracted from FIA Race Control data. Section 4 describes the experimental protocol, including the baseline detector and the diagnostics run to check conformal prediction's underlying assumptions. Section 5 reports results. Section 6 discusses why the method underperforms and what that implies. Section 7 states limitations plainly. Section 8 outlines what would need to change for this class of method to become viable.

## 2. Theoretical Framework

### 2.1 The Problem of Measuring Friction Directly

The foundational quantity in any traction-loss model is the tire-road friction coefficient, $\mu$. It is not directly observable in a standard telemetry stream , it requires specialized optical or acoustic sensors not deployed on the cars whose data is public, and it is highly sensitive to surface variation, temperature, and compound chemistry. This paper, like its predecessor draft, sidesteps $\mu$ by engineering a proxy from quantities available in public F1 telemetry: engine $RPM$, ground speed, throttle position, brake pressure, and gear selection.

### 2.2 Longitudinal Slip Proxy ($S_{proxy}$)

Under steady-state conditions in a mechanically coupled drivetrain, the ratio of engine $RPM$ to ground speed is approximately constant within a fixed gear. Deviations from that ratio indicate the driven wheels are rotating faster than the vehicle is travelling , the kinematic signature of longitudinal slip. The proxy is unchanged from the original formulation:

$$S_{proxy} = RPM/(v + \epsilon), \ \epsilon = 1 \times 10^{-5}$$

where $v$ is ground speed and $\epsilon$ prevents division instability during telemetry dropouts.

This proxy is dominated by two things: gear ratio and speed. A prior formulation of this work modeled $S_{proxy}$ using only throttle, brake pressure, and gear as predictors , leaving speed variation within a gear entirely unmodeled, even though speed is the denominator of the target. Because corner-exit acceleration profiles produce large, entirely non-slip speed changes within a fixed gear, that formulation risked attributing ordinary acceleration to “anomalous” behavior. This is corrected here: speed is included as an explicit fourth model feature, so that residuals reflect variation the model could not explain given throttle, brake, gear, and speed jointly , a materially different and more defensible quantity.

### 2.3 Inductive Conformal Calibration

For each driver, a Random Forest Regressor (100 $n_{estimators}$, unless swept in the sensitivity analysis) is trained on the calibration split to predict $S_{proxy}$ from throttle, brake pressure, gear, and speed. The non-conformity score for calibration sample $i$ is the absolute residual $\alpha_i = |y_i - \hat{\mu}(x_i)|$. The Calibration Bound $\hat{q}_{95}$ is the 95th percentile of these residuals. Unlike the original formulation, $\hat{q}_{95}$ is reported here with a block-bootstrap 95% confidence interval (500 resamples, block length 25 samples) rather than as a bare point estimate, to reflect sampling uncertainty and to partially respect the temporal dependence structure of the data (see Section 4.3).

### 2.4 Rolling Uncertainty Volatility ($\sigma_{UQ}$)

The Rolling Uncertainty Volatility metric is the moving standard deviation of non-conformity scores over a sliding window of width $k$. When $\sigma_{UQ}$ exceeds $\hat{q}_{95}$, the sample is flagged as a candidate pre-incident anomaly. This definition is unchanged from the original formulation; what changes in this paper is that flags are no longer interpreted narratively , every flag is checked against a timestamped, independently sourced incident record (Section 3), and scored as a true or false positive accordingly.

## 3. Ground-Truth Incident Extraction

A central gap in the original formulation of this work was the absence of any ground truth: flagged anomalies were interpreted through post-hoc narrative (e.g., attributing a driver's residual pattern to “aggressive defensive strategy”) rather than checked against anything that actually happened on track. This paper closes that gap, with the caveat that the available ground truth is narrower than would be ideal.

Two sources were used. First, FIA Race Control Messages for the session were scanned for entries matching flag states (yellow, double-yellow, red) or free-text mentions of spins, off-track excursions, incidents, or lock-ups, each associated with a session timestamp and, where available, a car number. Second, laps deleted for track-limits violations were extracted directly from the session's lap data, since a deleted lap indicates the car exceeded track boundaries at some point in that lap , a weak but independently verifiable proxy for a car being at or beyond the limit of control.

For the 2023 Monza race specifically, this extraction produced 14 timestamped incidents in total, and , this is itself a finding worth stating plainly , every one of them was a track-limits deletion. No yellow flag, spin, or lock-up messages matching the search criteria were logged by Race Control during this race. This is consistent with Monza 2023 having been a comparatively incident-free race at the front of the field, but it means the ground truth available for this evaluation is narrower than the paper's original framing implied: “ran wide at a corner” is a related but distinct event from “lost traction,” and the two should not be assumed interchangeable. Incidents were sparse , 0 to 3 per driver , which limits the statistical power of any precision/recall estimate computed per driver, a point we return to in Section 6.

## 4. Experimental Design

### 4.1 Data Scope

Telemetry was extracted using the FastF1 Python library for the full 2023 Italian Grand Prix race session. Unlike the original formulation, which used only each driver's single fastest lap, this evaluation concatenates telemetry across every racing lap for each driver (pit in/out laps excluded), so that laps containing genuine instability are not systematically excluded by construction. The same traction-zone filter as the original work is retained: samples are kept only where the engaged gear is 2, 3, or 4 and speed exceeds 30 km/h, isolating the power-application phases at the Ascari Chicane and Parabolica exits. Drivers with fewer than 100 usable samples after filtering were excluded; 19 of 20 classified drivers had sufficient data.

### 4.2 Chronological Split and Model Fitting

For each driver, the filtered, concatenated telemetry is split chronologically across the whole race , 50% calibration, 50% test , rather than within a single lap. A Random Forest Regressor is fit on the calibration split using throttle, brake, gear, and speed as features (Section 2.2). Non-conformity scores and $\hat{q}_{95}$ are computed on the calibration split; rolling volatility and flags are computed on the test split.

### 4.3 Autocorrelation Diagnostic

Split conformal prediction's coverage guarantee relies on exchangeability between calibration and test non-conformity scores. Telemetry sampled at high frequency is strongly serially correlated, which threatens that assumption. This was tested directly: a Ljung-Box test (lag 10) was run on the test-phase residuals for every driver. Results are reported in Section 5.3.

### 4.4 Baseline Detector

To evaluate whether the added complexity of rolling volatility monitoring is justified, a static baseline was run on identical data: a sample is flagged if its non-conformity score exceeds the calibration-split 95th-percentile residual directly, with no rolling window or volatility computation. Both detectors were scored identically against the same incident labels.

### 4.5 Detection Scoring

A flag is scored as a true positive if a ground-truth incident for that driver occurs within a 12-second window following the flag (flags must precede or coincide with the incident to carry predictive value; precision = matched flags ÷ total flags). An incident is scored as detected (for recall) if any flag occurred within a lookback window ending at the incident timestamp (recall = matched incidents ÷ total incidents). Mean and median lead time are reported in seconds where a match occurred

### 4.6 Sensitivity Analysis

To rule out the possibility that poor performance was an artifact of specific hyperparameter choices, both the rolling window size $k \in \{5, 10, 20, 50\}$ and the forest size $n_{estimators} \in \{25, 50, 100, 200\}$ were swept independently, holding the other fixed at its default ($k = 20$, $n_{estimators} = 100$), for every driver.

## 5. Results

### 5.1 Calibration Bounds

Calibration bounds ($\hat{q}_{95}$) ranged from 2.3375 (PIA) to 3.8105 (LAW) across the grid, computed with speed now included as a model feature. Block-bootstrap 95% confidence intervals are reported alongside each point estimate; note that these intervals frequently overlap across adjacent drivers, meaning the driver by-driver ranking in the calibration bound alone should be treated cautiously rather than as a precise ordering of mechanical stability.

| **Driver** | $\hat{q}_{95}$ **(naive)** | **Block-bootstrap 95% CI** | **Test samples** | **GT incidents (matched pool)** |
|---|---|---|---|---|
| PIA | 2.3375 | [2.1943, 2.5592] | 3063 | 1 |
| NOR | 2.4405 | [2.1789, 2.6043] | 4496 | 2 |
| RUS | 2.6881 | [2.4938, 2.9033] | 2963 | 1 |
| ALB | 2.7576 | [2.5794, 3.047] | 2579 | 0 |
| HAM | 2.8357 | [2.5805, 3.1114] | 2359 | 2 |
| STR | 2.8479 | [2.6462, 3.0539] | 3487 | 0 |
| ALO | 2.9395 | [2.686, 3.1432] | 3345 | 1 |
| SAI | 2.959 | [2.7465, 3.1469] | 2361 | 1 |
| HUL | 3.0325 | [2.8309, 3.4247] | 2722 | 0 |
| ZHO | 3.0882 | [2.8985, 3.2925] | 2433 | 0 |

| Driver | $\hat{q}_{95}$ (naive) | Block-bootstrap 95% CI | Test samples | GT incidents (matched pool) |
|---|---|---|---|---|
| BOT | 3.1458 | [2.918, 3.4593] | 2618 | 1 |
| LEC | 3.1576 | [2.8687, 3.4555] | 2984 | 1 |
| OCO | 3.173 | [2.6802, 3.6641] | 2801 | 0 |
| GAS | 3.2008 | [2.888, 3.5147] | 3789 | 0 |
| PER | 3.2316 | [2.9577, 3.5] | 3090 | 3 |
| VER | 3.3339 | [3.0607, 3.7493] | 2566 | 0 |
| MAG | 3.4005 | [3.0364, 3.6548] | 2410 | 0 |
| SAR | 3.4226 | [3.2092, 3.8313] | 2784 | 1 |
| LAW | 3.8105 | [3.4954, 4.1965] | 2713 | 0 |

## 5.2 Detection Performance: Conformal Volatility vs. Static Baseline

Table 2 aggregates detection performance across all 19 drivers with usable data, scored against the 14 ground-truth incidents extracted in Section 3.

| Method | Total flags (all drivers) | Mean flag rate | Mean precision | Mean recall |
|---|---|---|---|---|
| Rolling conformal volatility (proposed) | 8673 | 15.3% | 0.00000 | 0.000 |
| Static 95th-percentile threshold (baseline) | 13023 | 23.4% | 0.00014 | 0.050 |

Both methods perform poorly. The rolling conformal volatility detector produced 8,673 flags across all drivers and matched essentially none of the 14 ground-truth incidents (mean precision and recall both round to 0.0). The static threshold baseline produced more flags (13,023) and achieved one match , a single driver (Norris) for whom one flag preceded one deleted-lap incident by roughly 7.3 seconds , which is enough to move the aggregate recall figure to a non-zero but still negligible 0.05, driven entirely by that single event out of 14 total incidents. Precision remains near zero for the baseline as well (0.00014), because both methods flag such a large fraction of all samples that any single match is essentially guaranteed to occur eventually by chance, without indicating a reliable signal.

The flag rate itself is diagnostic. The conformal method flags a mean of 15.3% of all test samples (range

7.5%–31.7%) as anomalous; the baseline flags a mean of 23.4% (range 19.4%–28.5%). Neither rate is consistent with a signal meant to identify a rare, specific pre-incident state , a detector that fires on roughly one in every four to seven samples cannot function as an early-warning system regardless of what fraction of those flags happen to precede a genuine incident.

### 5.3 Autocorrelation Diagnostic

The Ljung-Box test at lag 10 rejected the null hypothesis of no autocorrelation for all 19 of 19 drivers ($p < 0.001$ in every case, and $p < 10^{-90}$ for the majority). Split conformal prediction's validity guarantee assumes exchangeable calibration/test residuals; this result shows that assumption does not hold for this data. This is a plausible, though not the only, contributor to the excessive flag rate observed above: strongly autocorrelated residuals violate the i.i.d.-style assumptions underlying the naive quantile-based calibration bound, and can produce runs of consecutive flagged samples that inflate the apparent flag rate without corresponding to independent anomalous events.

### 5.4 Sensitivity Analysis

Varying the rolling window size $k$ did not materially change the outcome. Mean precision across drivers stayed at or extremely close to zero for every value of $k$ tested:

| Window size | Mean precision across drivers | Mean flags per driver |
|---|---|---|
| 5 | 0.00032 | 340 |
| 10 | 0.00013 | 418 |
| 20 | 0.00000 | 456 |
| 50 | 0.00000 | 429 |

Varying the number of trees in the Random Forest likewise had no meaningful effect on detection performance:

| Window size | Mean precision across drivers | Mean flags per driver |
|---|---|---|
| 25 | 0.00000 | 453 |
| 50 | 0.00000 | 459 |
| 100 | 0.00000 | 456 |
| 200 | 0.00000 | 468 |

This is an important negative control: the failure to detect incidents is not attributable to a poorly chosen window size or an undertrained model. The signal that would need to exist for either hyperparameter to matter does not appear to be present in the residuals at all.

## 6. Discussion

Three explanations for these results are worth separating, because they carry different implications for whether this class of method could still work with modifications.

**First, the ground-truth definition is narrower than the paper's premise.** All 14 available incidents are track-limits deletions, not confirmed traction-loss events. A car can run wide for reasons unrelated to slip (a defensive line, a misjudged apex), and a car can lose traction without running wide enough to be flagged. Scoring against this label set measures something correlated with, but not identical to, the paper's actual target. It is possible , though untested here , that the method performs differently against a stricter or richer incident definition (e.g., telemetry-derived spin detection from yaw rate or lateral acceleration, which is not present in the public FastF1 dataset used).

**Second, the false-alarm rate is high enough to make the question partly moot.** A detector flagging 15– 32% of all samples cannot be operationally useful as a pre-incident warning regardless of how its precision/recall against a specific label set turns out, because a system that fires that often provides no discriminating information to a driver or controller. This is arguably the more important finding than the precision/recall numbers themselves: the method is not selective enough to be a warning signal in its current form.

**Third, the conformal validity assumption is violated.** Every driver's residuals show significant autocorrelation. This does not necessarily invalidate the general idea of residual-volatility monitoring, but it does mean the specific claim to statistical rigor , that $\hat{q}_{95}$ is a distribution-free, validly calibrated bound , is not actually true for this data as analyzed. Methods designed for dependent data (e.g., weighted or adaptive conformal prediction for time series) would need to replace the naive split conformal calibration used here before that claim could be made honestly.

Taken together, these results do not establish that residual-volatility monitoring of slip proxies can never work; they establish that the specific, naive implementation tested here , a plain 95th-percentile split conformal bound on autocorrelated residuals, evaluated against track-limits deletions as ground truth , does not produce a usable pre-incident detector on this race's data.

## 7. Limitations

Beyond the diagnostics above, several limitations bound how far these results should be generalized.

**Single race, single session.** All results are from one race (2023 Italian Grand Prix). Track characteristics, weather, and race dynamics vary substantially across a season; nothing here should be read as a claim about F1 telemetry monitoring in general.

**The slip proxy remains a kinematic approximation, not a direct measurement.** Even with speed included as a feature, S_proxy cannot distinguish driven-axle spin from tire deformation under cornering load without lateral acceleration data, which is not available in the telemetry channels used.

**Ground truth is incident-sparse by nature.** With only 14 incidents across the full grid, per-driver precision and recall estimates are built on very small denominators (0–3 incidents per driver) and should not be over-interpreted individually; the aggregate picture across drivers is more informative than any single driver's row in Table 2.

**The 12-second match window and 6-second recall lookback were fixed choices, not tuned or validated against alternative values.** A more exhaustive study would sweep these as well.

**Deployment latency was not evaluated.** Even a correctly calibrated version of this method would need to run at 50–100 Hz on embedded hardware for real-time use; that engineering question is untouched by this paper.

## 8. What a Viable Version Would Require

Based on the diagnosis in Section 6, a version of this idea with a real chance of working would need, at minimum: (1) a richer ground-truth label , ideally derived from a physical quantity such as lateral/longitudinal acceleration spikes or yaw-rate discontinuities rather than track-limits deletions alone; (2) a conformal calibration method designed for temporally dependent data (e.g., NexCP-style or adaptive conformal inference approaches that explicitly handle non-exchangeable, autocorrelated sequences) rather than naive split conformal; (3) evaluation across many races and drivers to obtain enough true incidents for precision/recall estimates with usable statistical power , 14 incidents from one race is not enough to confidently distinguish a working detector from noise; and (4) an explicit false alarm-rate constraint built into the flagging rule itself, since an unconstrained 15–30% flag rate is disqualifying on its own regardless of what ground truth is used.

## 9. Conclusion

This paper set out to evaluate, rather than assume, whether rolling split conformal prediction on F1 telemetry residuals can serve as a pre-incident traction-loss warning signal. After correcting a proxy confound present in an earlier formulation, using full-race data instead of only each driver's cleanest lap, and scoring flagged events against real, independently sourced incident timestamps, the method does not demonstrate the capability the underlying idea promised: precision and recall against ground truth are effectively zero, the false-alarm rate is too high for operational use, and the statistical assumption conformal prediction relies on is empirically violated in this data. We report this as the honest outcome of a rigorous test, along with a concrete diagnosis of the three most likely causes and what would need to change for a future version of this approach to be evaluated fairly.

## Appendix A: Data and Code Availability

The corrected analysis pipeline (proxy construction, ground-truth extraction from Race Control Messages, chronological split, block-bootstrap calibration, autocorrelation diagnostics, baseline detector, and sensitivity sweep) and the full CSV outputs underlying every table in this paper are available alongside this manuscript for independent verification. Z

https://github.com/nearpot/predictive-conformal-slip-monitoring